\documentclass[preprint,NumberedRefs]{JASA}
\usepackage{siunitx}
\usepackage{booktabs}
\usepackage{multirow}
\usepackage{needspace}

\let\linenumbers\nolinenumbers

\hypersetup{pdftitle={Compressed delayed-information projection for underwater navigation under acoustic measurement latency}}
\begin{document}

\preprint{Submitted to The Journal of the Acoustical Society of America}

\title[]{Compressed delayed-information projection for underwater navigation under acoustic measurement latency}

\author{Shuyue Li}
\affiliation{Academy of Artificial Intelligence and Advanced Technology, Xi'an Jiaotong-Liverpool University, Suzhou, Jiangsu 215123, China}
\affiliation{Department of Electrical Engineering and Electronics, School of Engineering, University of Liverpool, Liverpool L69 3GJ, United Kingdom}
\affiliation{Jiangsu JITRI Tsingunited Intelligent Control Technology Co., Ltd., Building D1, No. 999 Gaolang East Road, Wuxi, Jiangsu 214131, China}
\author{Miguel L\'opez-Ben\'itez}
\affiliation{School of Computer Science and Informatics, University of Liverpool, Liverpool L69 3DR, United Kingdom}
\affiliation{ARIES Research Centre, Polytechnic School, Universidad Antonio de Nebrija, Madrid 28040, Spain}
\author{Eng Gee Lim}
\affiliation{Academy of Artificial Intelligence and Advanced Technology, Xi'an Jiaotong-Liverpool University, Suzhou, Jiangsu 215123, China}
\author{Fei Ma}
\affiliation{School of Mathematics and Physics, Xi'an Jiaotong-Liverpool University, Suzhou, Jiangsu 215123, China}
\author{Qian Dong}
\affiliation{Academy of Artificial Intelligence and Advanced Technology, Xi'an Jiaotong-Liverpool University, Suzhou, Jiangsu 215123, China}
\author{Mengze Cao}
\affiliation{Jiangsu JITRI Tsingunited Intelligent Control Technology Co., Ltd., Building D1, No. 999 Gaolang East Road, Wuxi, Jiangsu 214131, China}
\affiliation{College of Mechanical and Vehicle Engineering, Hunan University, Changsha, Hunan 410082, China}
\author{Limin Yu}
\email{limin.yu@xjtlu.edu.cn}
\affiliation{Academy of Artificial Intelligence and Advanced Technology, Xi'an Jiaotong-Liverpool University, Suzhou, Jiangsu 215123, China}
\author{Xiaohui Qin}
\email{qinxiaohui@tsingunited.com}
\affiliation{Jiangsu JITRI Tsingunited Intelligent Control Technology Co., Ltd., Building D1, No. 999 Gaolang East Road, Wuxi, Jiangsu 214131, China}

\begin{abstract}
Delayed acoustic position measurements describe past states but may reach an underwater vehicle after its navigation estimate has advanced. This paper presents compressed delayed-information projection (CDIP), a causal delayed-update method implemented in a 15-dimensional error-state Kalman filter (ESKF). CDIP retains source-associated inertial measurement unit (IMU) snapshots and maintains current-to-snapshot cross-covariances, projecting delayed corrections without repeated full-from-start filtering. In 154 paired simulation recordings with synthetic horizontal-position fixes delayed by 1.5~s, CDIP reduced mean trajectory-position root-mean-square error (RMSE) from 1.062~m for a reception-time Baseline to 0.456~m (57.1\%). Its mean RMSE approached that of full-from-start Replay (0.451~m), with 99.2\% lower mean core-loop time per processed IMU step for the tested implementations. A six-delay sweep, scheduled permanent-loss outages of 50--300~s, and packet-erasure tests further characterized sensitivity to measurement timing and availability. A separate 256-trial nonlinear synthetic assessment identified departures from the prespecified calibration criterion for CDIP and two fully augmented ESKF references. Together, these results demonstrate an accuracy--computation trade-off for causal underwater navigation under the evaluated delayed-measurement simulation conditions.
\end{abstract}

\maketitle

\section{Introduction}
\label{sec:introduction}

Unmanned underwater vehicles (UUVs), also termed autonomous underwater vehicles (AUVs), require navigation for mapping, monitoring and inspection \cite{Kinsey2006IFAC,Paull2014,Maurelli2022AUV}. Underwater sensor-network monitoring also requires location information \cite{Chowdhury2025NextGen}. Acoustic links have limited bandwidth and long propagation delays \cite{Heidemann2012USN,Chitre2008}; time-varying multipath complicates acoustic equalization \cite{Zhang2026Turbo}. An editorial discusses underwater communication, navigation and localization \cite{Ullah2025}. Channel studies document variability, multipath and equalization effects \cite{Yang2012Channel,Preisig2005TimeVarying,Rouseff2009Kauai}, while spatial diversity has been investigated to mitigate interference-induced message loss \cite{McGee2014SpatialDiversity}. Current research spans underwater acoustic medium-access control \cite{Huangfu2026PPMAC} and asynchronous AUV communication/positioning networks \cite{Wang2025Asynchronous}. A survey discusses cooperative UUV estimation and fusion \cite{Li2025Drones}. Passive acoustic monitoring with AUV teams has been studied under intermittent communication \cite{Tiranti2025Motion}, and cooperative navigation filters address acoustic delays \cite{li2026asynchronous}. A received acoustic packet can therefore describe a state earlier than the current filter estimate.

Strapdown inertial navigation systems (INS) propagate motion estimates using inertial measurement units (IMUs) \cite{Titterton2004Book}. Underwater navigation systems combine INS with Doppler velocity logs (DVLs), depth sensing and acoustic aiding \cite{Kinsey2006IFAC,Paull2014}. Acoustic aiding includes travel-time-based AUV navigation \cite{Wang2025OneWay}, while passive source localization can exploit intensity striations \cite{Turgut2010SourceLocalization}. A review describes operational sensor choices for underwater localization \cite{Sorensen2025Localization}. Inertial dead reckoning accumulates errors and benefits from external observations \cite{Zhang2023LocalizationUncertainty,Tang2024UnderwaterError}. Aiding measurements can arrive late, motivating explicit delay treatment \cite{Li2024DelayAUV,Chen2023Treatment,Xu2021Novel,Yan2026Delay}.

Out-of-sequence measurement (OOSM) updates and moving-horizon estimation treat delayed measurements \cite{BarShalom2004OOSM,Gao2015MHE}. Acoustic measurement-equation reconstruction provides another EKF-based approach \cite{Li2024DelayAUV}. Stochastic cloning retains temporal correlations for relative-state measurements \cite{RoumeliotisBurdick2002Cloning,Mourikis2007SCKF}; classical exact OOSM updates address tracking \cite{BarShalom2002Exact}. Related cooperative and distributed localization methods provide precedents \cite{Roumeliotis2002,Bahr2009Cooperative}. More recent multi-AUV methods address delay-biased observations and unreliable inputs \cite{Wang2025Enhanced}. General estimation and delayed-navigation treatments provide context \cite{BarShalom2001,Frei2023Robust}. These studies provide context for the present implementation-specific comparison of reception-time Baseline, CDIP and full-from-start Replay.

For this OOSM problem \cite{Fossen2023Delay}, Baseline evaluates a historical observation against the reception-time state. Replay instead reruns the causal filter with then-received packets scheduled at their source-associated IMU epochs.

Established OOSM methods treat delayed measurements \cite{BarShalom2002Exact,BarShalom2004OOSM}. The authors' OCEANS variational history distillation (VHD) synthesizes outage-time virtual measurements \cite{li2026communicationoutageresistantuuvstate}; their asynchronous two-speed preprint combines Gaussian-process (GP) prediction and buffered transition-product projection for cooperative navigation \cite{li2026asynchronous}. CDIP instead maintains 15-state current-to-source blocks through causal updates and error resets for a single UUV. The earlier systems use different states and aiding protocols; the present comparison uses the single-UUV configuration defined below.

This estimator-level study uses recorded vehicle inputs and synthetic horizontal acoustic-position fixes. CDIP stores post-DVL IMU-grid snapshots and projects delayed corrections through maintained cross-covariances. Its frozen-history representation provides an approximate connection between observations generated at an earlier time and the current-state correction.

The contributions are as follows:
\begin{itemize}
  \item \textbf{Method:} A causal 15-state ESKF implementation with a frozen-history projection retaining source snapshots and current-to-source covariance through prediction, onboard updates and right-error reset. This representation differs from VHD virtual measurements and two-speed transition products \cite{li2026communicationoutageresistantuuvstate,li2026asynchronous}.
  \item \textbf{Implementation trade-off:} Delayed corrections avoid repeated IMU, DVL, depth, reset and acoustic passes. The full-from-start Replay ESKF supplies the accuracy reference and the implementation-specific timing comparison.
  \item \textbf{Controlled evaluation:} A comparison using identical inputs and deterministic realizations across three treatments in 154 paired recordings at a fixed delay of 1.5~s without outage; a predeclared six-delay sweep with 30 pairs/delay and a zero-delay agreement check; descriptive 9-D covariance diagnostics; separate scheduled permanent-loss outages lasting 50--300~s with five pairs/duration; and random packet-erasure sensitivity.
\end{itemize}

The remainder of this paper is organized as follows.
Section~\ref{sec:problem} introduces the navigation and delayed-acoustic measurement models.
Sections~\ref{sec:method} and~\ref{sec:analysis} present the CDIP formulation, the Baseline and Replay reference methods, and the error-state implementation.
Section~\ref{sec:experiments} reports the primary fixed-delay experiment, the separate six-delay sweep, targeted outage tests, and synthetic packet-erasure sensitivity.
The Discussion and Conclusion interpret the findings, limitations, and future research directions.

\section{System Model and Problem Formulation}
\label{sec:problem}

This section defines the system and measurement models for a six-degree-of-freedom (6-DOF) underwater vehicle with a causal filter and formulates the delayed-acoustic navigation problem.

\Needspace{12\baselineskip}
\subsection{System state definition}
\label{subsec:state}

The navigation state of the underwater vehicle is defined in a global navigation frame as
\begin{equation}
\mathbf{x}_k =
\begin{bmatrix}
\mathbf{p}_k \\
\mathbf{v}_k \\
\mathbf{q}_k \\
  \mathbf{b}^\omega_k \\
  \mathbf{b}^a_k
\end{bmatrix}
\in \mathbb{R}^{16},
\end{equation}
where $\mathbf{p}_k \in \mathbb{R}^3$ and $\mathbf{v}_k \in \mathbb{R}^3$ denote the vehicle position and velocity in the navigation frame, respectively, $\mathbf{q}_k \in \mathbb{R}^4$ represents the unit quaternion describing vehicle attitude, and $\mathbf{b}^\omega_k \in \mathbb{R}^3$ and $\mathbf{b}^a_k \in \mathbb{R}^3$ are the gyroscope and accelerometer bias vectors.

The unit quaternion has four nominal components, while its small attitude error is represented by three tangent-space coordinates on $SO(3)$. This gives a 16-component nominal state and a 15-dimensional error state.

\subsection{Motion model}
\label{subsec:motion}

Standard strapdown navigation provides the propagation framework \cite{Titterton2004Book}. Marine-craft kinematics and integrated navigation are treated by Fossen \cite{Fossen2011Handbook}. The continuous-time propagation model used here is:
\begin{equation}
\begin{aligned}
\dot{\mathbf{p}} &= \mathbf{v}, \\
\dot{\mathbf{v}} &= \mathbf{R}(\mathbf{q})\left(\mathbf{a}_m - \mathbf{b}^a - \mathbf{n}_a\right) + \mathbf{g}, \\
\dot{\mathbf{q}} &= \frac{1}{2}\boldsymbol{\Omega}\left(\boldsymbol{\omega}_m - \mathbf{b}^\omega - \mathbf{n}_\omega\right)\mathbf{q}, \\
\dot{\mathbf{b}}^a &= \mathbf{n}_{ba}, \\
\dot{\mathbf{b}}^\omega &= \mathbf{n}_{b\omega},
\end{aligned}
\end{equation}
Here $\mathbf{a}_m$ and $\boldsymbol{\omega}_m$ are recorded inputs interpreted by the estimator as body-frame specific force and angular velocity; $\mathbf{R}(\mathbf{q})$ maps body to navigation coordinates and is also denoted $\mathbf{R}_{nb}$; $\boldsymbol{\Omega}$ is the quaternion angular-rate matrix. Positive z is upward; signed gravity is $\mathbf{g}=-(0,0,g_0)^\top$, with $g_0=9.80665\,\mathrm{m\,s^{-2}}$. Propagation subtracts the stored positive-z reference $-\mathbf{g}$; IMU values receive no additional noise draws. An independent Seed~1 raw-IMU diagnostic supports the body-frame gyroscope interpretation, but insufficient three-axis excitation leaves the full mounting rotation unidentified. The recorded frame is \texttt{rexrov/imu\_link}; its mounting rotation and lever arm, physical bias, and production noise/seed/reset configuration remain unverified. The process and bias equations specify estimator assumptions.

\subsection{Measurement models}
\label{subsec:measurement}

The underwater vehicle is equipped with onboard aiding sensors, including a DVL, which measures three-dimensional vehicle-frame velocity, and a depth sensor, which provides vertical position \cite{Jeong2024SensorFusion}. These measurements are locally available at their respective sampling times. Let $\mathbf{R}_{bd}$ map a DVL-link vector into the body frame, $\mathbf{R}_{db}=\mathbf{R}_{bd}^{\top}$, and $\mathbf{R}_{bn}=\mathbf{R}_{nb}^{\top}$. For the Gazebo-simulated RexROV UUV model, the estimator uses the DVL extrinsic
\begin{equation}
\mathbf{p}_{bd} = [-1.4,\;0,\;-0.312]^{\top}\ \mathrm{m},
\qquad
\mathbf{R}_{bd}=\mathbf{R}_y(\pi/2).
\end{equation}
The DVL prediction includes both translational and lever-arm rotational velocity:
\begin{equation}
\mathbf{z}^{\mathrm{DVL}}_k
=
\mathbf{R}_{db}\left(\mathbf{R}_{bn}\mathbf{v}_k+
\boldsymbol{\omega}^b_k\times\mathbf{p}_{bd}\right)
+
\mathbf{v}^{\mathrm{DVL}}_k,
\end{equation}
where the body angular rate entering the lever-arm term is the bias-corrected IMU angular-rate measurement,
\begin{equation}
\boldsymbol{\omega}^b_k = \boldsymbol{\omega}_{\mathrm{IMU},k}-\mathbf{b}^{\omega}_k.
\end{equation}
The IMU angular rate is bias-corrected before the DVL lever-arm prediction.
\begin{equation}
\mathbf{z}^{\mathrm{depth}}_k
=
\mathbf{H}_{\mathrm{depth}}\mathbf{x}_k
+
\mathbf{v}^{\mathrm{depth}}_k,
\end{equation}
where the DVL part of the right-error Jacobian has velocity, attitude, and gyroscope-bias blocks $\mathbf{R}_{db}\mathbf{R}_{bn}$, $\mathbf{R}_{db}[\mathbf{R}_{bn}\mathbf{v}_k]_{\times}$, and $\mathbf{R}_{db}[\mathbf{p}_{bd}]_{\times}$, respectively. The depth measurement is $z_k^{\mathrm{depth}}=p_{z,k}+v_k^{\mathrm{depth}}$.

In addition to the onboard measurements, the UUV receives acoustic positioning updates from an external underwater acoustic reference. In the present study, these acoustic observations provide horizontal two-dimensional (2-D) position information only. For packet $j$ generated at its actual DVL/acoustic timestamp $\tau_j$, the acoustic observation is modeled as
\begin{equation}
\mathbf{z}^{\mathrm{ac}}_j
=
\mathbf{H}_{\mathrm{ac}}^{\mathrm{nom}}\mathbf{x}(\tau_j)
+
\mathbf{v}^{\mathrm{ac}}_j,
\end{equation}
where $\mathbf{H}_{\mathrm{ac}}^{\mathrm{nom}}=[\mathbf{I}_2\ \mathbf{0}_{2\times14}]$ selects horizontal position from the 16-component nominal state and $\mathbf{v}^{\mathrm{ac}}_j \sim \mathcal{N}(\mathbf{0},\mathbf{R}_{\mathrm{ac}})$. The corresponding 15-state error Jacobian is $\mathbf{H}_{\mathrm{ac}}=[\mathbf{I}_2\ \mathbf{0}_{2\times13}]$. The evaluated covariance is fixed at $\mathbf{R}_{\mathrm{ac}}=25\mathbf{I}_{2}$~m$^2$, corresponding to a 5-m standard deviation per horizontal axis; it is not derived from received signal-to-noise ratio.

The recorded DVL signal is preserved as a raw input. The final experiments apply the declared deterministic offline Gaussian DVL measurement-noise realization ($\sigma_{\mathrm{DVL}}=0.05$~m/s); this is distinct from raw Gazebo DVL noise. Depth and horizontal acoustic observations are explicitly generated from ground truth at DVL time stamps under the declared measurement policy.

\subsection{Delayed acoustic observation model}
\label{subsec:delay_model}

Underwater acoustic information exchange is affected by finite sound-propagation speed and limited bandwidth. These effects cause acoustic positioning observations to become available to the navigation estimator asynchronously. Let packet $j$ be generated at $\tau_j$ and processed on reception at IMU time $t_k\geq\tau_j$. Let $s(j)$ denote the IMU index at which its associated DVL update is processed and the post-DVL, pre-acoustic snapshot is stored, with $t_{s,j}=t_{s(j)}\geq\tau_j$ and $\delta_j=t_{s,j}-\tau_j$. Thus, the physical generation time and stored-snapshot time are distinct. Its age upon reception is
\begin{equation}
a_{k,j}
=
t_k-\tau_j,
\end{equation}
whereas its processing-grid separation is
\begin{equation}
d_{k,j}
=
k-s(j),
\end{equation}
which need not equal the physical age divided by an IMU sampling interval. The received acoustic observation still measures the state at $\tau_j$:
\begin{equation}
\mathbf{z}^{\mathrm{ac}}_{j \rightarrow k}
=
\mathbf{H}_{\mathrm{ac}}^{\mathrm{nom}}\mathbf{x}(\tau_j)
+
\mathbf{v}^{\mathrm{ac}}_{j},
\qquad \tau_j\leq t_k.
\end{equation}

In a deployed system, acoustic observations can arrive over an underwater link after their generation epoch. A one-way propagation-only delay would satisfy $\tau=R/c$, whereas a transponder round trip would contribute $2R/c$ before processing and turnaround time; the present generator implements neither range $R$ nor sound speed $c$. Time-varying multipath and surface scattering affect acoustic equalization \cite{Preisig2005TimeVarying}. Propagation delay, clock synchronization and modem/network protocols affect underwater data delivery \cite{Chitre2008}.

\begin{figure}[t]
\centering
\includegraphics[width=8.5cm]{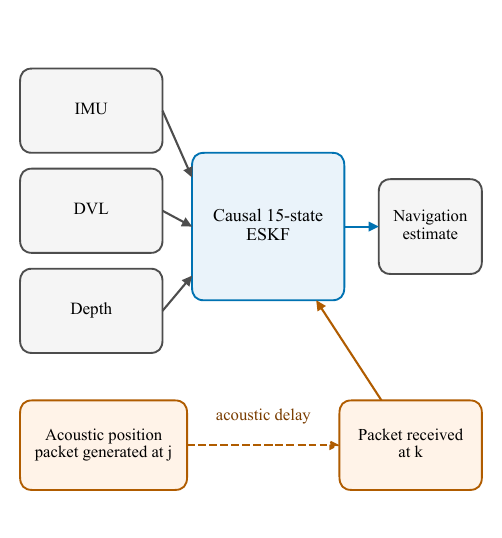}
\caption{Causal sensing and delayed-acoustic context. IMU, DVL, and depth observations are locally available to the causal ESKF. An acoustic position packet generated at $\tau_j$ becomes available at reception epoch $k$ and physically measures a historical state; the implemented source-associated snapshot is stored at the IMU processing time $t_{s,j}\geq\tau_j$.}
\label{fig:causal_context}
\end{figure}

Here the horizontal fix and its reception time are generated synthetically with fixed Gaussian covariance and an imposed delay; the targeted outage study imposes scheduled permanent loss, whereas the packet-erasure study uses fixed-seed random masks. The received observation consequently constrains the state at $\tau_j$, not at reception; the implementation associates it with the stored snapshot at $t_{s,j}$. Figure~\ref{fig:causal_context} summarizes this causal sensing and delayed-packet timing.

\subsection{Problem statement}
\label{subsec:problem_statement}

The objective is causal 6-DOF navigation with delayed acoustic fixes and onboard aiding. The three treatments share data and measurement realizations; all operate on the IMU grid, with CDIP using a source-associated snapshot rather than the exact generation state.

\section{Compressed Delayed-Information Projection}
\label{sec:method}

Baseline (reception-time update), CDIP and Replay (full-from-start ESKF rerun) share sensor models, initialization, covariance settings and deterministic realizations. Only delayed-packet processing differs.

\subsection{Compared delayed-acoustic treatments}
\label{subsec:overview}

The treatments of a received delayed acoustic packet are:
\begin{itemize}
  \item \textbf{Baseline:} the delayed packet is evaluated at reception using the current state and covariance.
  \item \textbf{CDIP:} the innovation against a source-associated IMU snapshot is projected to the current state through a maintained current-to-snapshot cross-covariance.
  \item \textbf{Replay:} at each reception, the common filter is rerun from initialization with the packets received so far scheduled at their source-associated IMU processing epochs.
\end{itemize}

\subsection{Baseline error-state filtering architecture}
\label{subsec:eskf}

At each DVL/depth epoch, the common filter performs an onboard update and right-error reset. Baseline applies a received acoustic innovation as if its source observation were current:
\begin{equation}
\mathbf{r}^{\mathrm{Baseline}}_{k,j}=\mathbf{z}^{\mathrm{ac}}_j-\hat{\mathbf{p}}_{xy,k}^{\mathrm{ac-}},
\end{equation}
where $\hat{\mathbf{p}}_{xy,k}^{\mathrm{ac-}}$ is the current horizontal estimate after any due onboard update and before the acoustic correction. Baseline uses the current covariance, without retaining the historical observation relation.

\subsection{Compressed delayed-information projection}
\label{subsec:cdip}

\begin{figure*}[t]
\centering
\includegraphics[width=17.0cm]{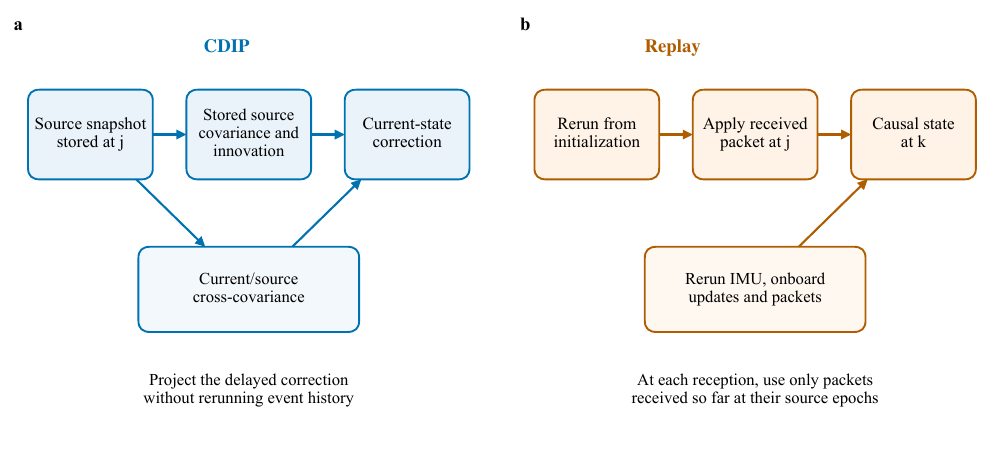}
\caption{CDIP and full-from-start Replay. (a) CDIP uses a source-associated snapshot stored at IMU processing time $t_{s,j}$ and maintained current-to-snapshot cross-covariance to project the delayed correction. (b) Replay reruns the common ESKF from initialization at each packet reception with only then-received acoustic packets scheduled at their source-associated IMU epochs, and takes the causal estimate at reception. The acoustic fix represents the earlier generation time $\tau_j$ in both paths; the Replay path supplies an accuracy reference through full-from-start reprocessing at source-associated IMU epochs.}
\label{fig:cdip_projection}
\end{figure*}

For each acoustic packet $j$, CDIP stores a post-DVL, pre-acoustic snapshot at $t_{s,j}$: its horizontal estimate $\hat{\mathbf{p}}_{xy,s,j}$ and 15-state error covariance $\mathbf{P}_j$. It maintains the $15\times15$ current-to-snapshot block $\mathbf{P}_{kj}$ through intervening predictions, onboard updates, and right-error resets. The stored position and $\mathbf{P}_j$ retain their snapshot values while the cross-covariances evolve with subsequent filtering. At reception, the implemented innovation and modeled covariance are
\begin{equation}
\mathbf{r}_{k,j}=\mathbf{z}^{\mathrm{ac}}_j-\hat{\mathbf{p}}_{xy,s,j},
\qquad
\mathbf{S}_{k,j}=\mathbf{H}_{\mathrm{ac}}\mathbf{P}_j\mathbf{H}_{\mathrm{ac}}^{\top}+\mathbf{R}_{\mathrm{ac}},
\label{eq:cdip_innovation}
\end{equation}
Let $\delta\mathbf{x}_{s,j}$ be the 15-component right-error state at $t_{s,j}$, whose position block is $\mathbf{p}(t_{s,j})-\hat{\mathbf{p}}_{s,j}$. The physical observation-time model then gives
\begin{equation}
\begin{aligned}
\mathbf{r}_{k,j}
&=\mathbf{H}_{\mathrm{ac}}\delta\mathbf{x}_{s,j}
  +\boldsymbol{\epsilon}^{\mathrm{time}}_j+\mathbf{v}^{\mathrm{ac}}_j,\\
\boldsymbol{\epsilon}^{\mathrm{time}}_j
&=\mathbf{p}_{xy}(\tau_j)-\mathbf{p}_{xy}(t_{s,j})
 =-\int_{\tau_j}^{t_{s,j}}\mathbf{v}_{xy}(t)\,\mathrm{d}t
\end{aligned}
\label{eq:source_time_offset}
\end{equation}
for a differentiable trajectory; the synthetic generator uses recorded ground-truth interpolation. The temporal term and frozen source error need not be conditionally centered. Modeled $\mathbf{S}_{k,j}$ omits temporal variance and source cross-covariance; the gain omits current-error cross-covariance. Stored blocks are not conditioned on all then-available observations, and $\mathbf{R}_{\mathrm{ac}}$ does not absorb these differences. The implemented projection is
\begin{equation}
\mathbf{K}_{kj}=\mathbf{P}_{kj}\mathbf{H}_{\mathrm{ac}}^{\top}\mathbf{S}_{k,j}^{-1},
\qquad
\delta\mathbf{x}_k=\mathbf{K}_{kj}\mathbf{r}_{k,j}.
\label{eq:cdip_gain}
\end{equation}
Before nominal-state injection and reset, the implemented current covariance is updated as
\begin{equation}
\widetilde{\mathbf{P}}_{k}=\mathbf{P}_{k}^{\mathrm{ac-}}-\mathbf{K}_{kj}\mathbf{S}_{k,j}\mathbf{K}_{kj}^{\top},
\end{equation}
followed by nominal-state injection and reset. Onboard updates map the current-to-snapshot block while retaining stored source marginals and source-pair blocks. Full joint conditioning would revise those historical statistics as well; CDIP therefore uses an approximate delayed-information representation. Approximation error can depend on intervening updates and assumed noise; its quantitative dependence and monotonicity require separate characterization. Section~\ref{subsec:complexity} gives its storage and scaling; Fig.~\ref{fig:cdip_projection} contrasts the treatments.

\subsection{Causal full-from-start Replay reference}
\label{subsec:replay}

At each acoustic reception, Replay reruns the common ESKF from initialization using only then-received packets, reassigning their reception times to generation timestamps. The IMU loop processes them at source-associated grid epochs, not continuously at $\tau_j$. The causal reception-time output comes from that full pass, without restoring a pre-source checkpoint.

\subsection{Algorithm summary}
\label{subsec:algorithm}

Algorithm~\ref{alg:cdip} summarizes the causal CDIP treatment of delayed acoustic packets.

\begin{algorithm}[t]
\caption{Compressed delayed-information projection}
\label{alg:cdip}
\begin{algorithmic}[1]
\STATE Initialize nominal state, current covariance, and an empty snapshot buffer
\FOR{each time step $k$}
\STATE Propagate nominal state using inertial measurements
\STATE Predict current covariance and map retained current-to-snapshot blocks
\IF{onboard measurements available}
\STATE Perform the DVL/depth update, error injection, and reset; map retained current-to-snapshot blocks
\STATE Store the post-DVL, pre-acoustic snapshot and initialize its covariance blocks
\ENDIF
\IF{one delayed acoustic packet is due}
\IF{the packet is available}
\STATE Retrieve its source-associated IMU snapshot and maintained $\mathbf{P}_{kj}$
\STATE Form $\mathbf{r}_{k,j}$ and $\mathbf{S}_{k,j}$ using Eq.~\eqref{eq:cdip_innovation}, then $\mathbf{K}_{kj}$ using Eq.~\eqref{eq:cdip_gain}
\STATE Update the current state, covariance, and maintained current-to-snapshot cross-covariances
\ENDIF
\STATE Remove its snapshot and advance the packet index
\ENDIF
\ENDFOR
\end{algorithmic}
\end{algorithm}

\section{Error-State Modeling and Implementation}
\label{sec:analysis}

\subsection{Error state and linearized dynamics}
\label{subsec:error_state}

The 15-dimensional error state is
\begin{equation}
\delta \mathbf{x} =
\begin{bmatrix}
\delta \mathbf{p}^\top &
\delta \mathbf{v}^\top &
\delta \boldsymbol{\theta}^\top &
\delta \mathbf{b}^{\omega\top} &
\delta \mathbf{b}^{a\top}
\end{bmatrix}^\top,
\end{equation}
The blocks are position, velocity, right-attitude, gyroscope-bias and accelerometer-bias errors. The quaternion is normalized after propagation and injection.

\label{subsec:error_dynamics}

Linearization around the nominal trajectory gives
\begin{equation}
\dot{\delta \mathbf{x}} = \mathbf{F}\delta \mathbf{x} + \mathbf{G}\mathbf{w},
\end{equation}
where $\mathbf{F}$ is the continuous-time error-dynamics Jacobian, $\mathbf{G}$ is the noise input matrix, and $\mathbf{w}$ denotes the stacked process noise vector.

\subsection{Covariance and retained historical blocks}
\label{subsec:covariance}

The error covariance $\mathbf{P}$ is propagated according to
\begin{equation}
\dot{\mathbf{P}} = \mathbf{F}\mathbf{P} + \mathbf{P}\mathbf{F}^\top + \mathbf{Q},
\end{equation}
where $\mathbf{Q}$ denotes the effective process-noise intensity in this covariance equation. For the three evaluated methods, the implementation sets $\mathbf{Q}=\mathbf{Q}_c=\operatorname{blkdiag}(\mathbf{0}_3,\allowbreak 10^{-6}\mathbf{I}_3,\allowbreak 10^{-8}\mathbf{I}_3,\allowbreak 10^{-10}\mathbf{I}_3,\allowbreak 10^{-10}\mathbf{I}_3)$ with the Gaussian-process correction disabled; the first-order discrete contribution is $\mathbf{Q}_k=\mathbf{Q}_c\Delta t$. The initial covariance $\mathbf{P}_0$ has five diagonal three-dimensional blocks formed from the squared configured position, velocity, attitude, gyroscope-bias, and accelerometer-bias standard deviations of $0.10$~m, $0.10$~m/s, $0.01$~rad, $0.01$~rad/s, and $0.01$~m/s$^2$, respectively. The measurement covariances are $\mathbf{R}_{\mathrm{DVL}}=(0.05)^2\mathbf{I}_3$~m$^2$/s$^2$, $R_{\mathrm{depth}}=(0.10)^2$~m$^2$, and $\mathbf{R}_{\mathrm{ac}}=(5.0)^2\mathbf{I}_2$~m$^2$. These intensities configure the filter; they are not measured IMU noise densities or a verified simulator noise/bias law.

For a retained source packet $i$, let $\mathbf{P}_{ki}$ denote the maintained $15\times15$ current-to-snapshot block and $\mathbf{B}_{ji}$ the stored $15\times15$ block between snapshots $j$ and $i$; $k$ indexes IMU processing, while $i,j$ index packets. At insertion at $k=s(i)$, $\mathbf{P}_{s(i),i}=\mathbf{B}_{ii}=\mathbf{P}_i$, and a newly stored $\mathbf{B}_{ji}$ is copied from the then-current $\mathbf{P}_{s(j),i}$. These blocks refer to $t_{s,i}$ and $t_{s,j}$, not exactly to $\tau_i$ and $\tau_j$. The implemented block maps are
\begin{align}
\mathbf{P}_{ki}^{\mathrm{pred}} &= \mathbf{\Phi}_k\mathbf{P}_{k-1,i}^{+}, \\
\mathbf{P}_{ki}^{\mathrm{on+}} &= \mathbf{J}_k^{\mathrm{on}}(\mathbf{I}-\mathbf{K}_k^{\mathrm{on}}\mathbf{H}_k^{\mathrm{on}})\mathbf{P}_{ki}^{\mathrm{pred}}, \\
\mathbf{P}_{ki}^{\mathrm{ac+}} &= \mathbf{J}_k^{\mathrm{ac}}\bigl(\mathbf{P}_{ki}^{\mathrm{ac-}}-\mathbf{K}_{kj}\mathbf{H}_{\mathrm{ac}}\mathbf{B}_{ji}\bigr),
\label{eq:implemented_cross_blocks}
\end{align}
The middle map applies to DVL/depth updates; $\mathbf{P}_{ki}^{\mathrm{ac-}}$ is the pre-acoustic block. With $\mathbf{H}_{\mathrm{ac}}$ of size $2\times15$ and $\mathbf{K}_{kj}$ of size $15\times2$, each map remains $15\times15$. The first-order right-error reset is $\mathbf{J}=\operatorname{blkdiag}(\mathbf{I}_6,\mathbf{I}_3-\tfrac12[\delta\boldsymbol{\theta}]_{\times},\mathbf{I}_6)$. After the gain in Eq.~\eqref{eq:cdip_gain}, current covariance is reset, symmetrized and eigenvalue-floored if needed. Stored source positions, marginals and $\mathbf{B}_{ji}$ retain their snapshot values. Equation~\eqref{eq:implemented_cross_blocks} specifies approximate cross-covariance propagation on the IMU grid as part of the frozen-history projection.

Under independent centered primitive errors and prescribed maps, frozen-history snapshot estimates admit an affine error-moment interpretation. Independent recomputation over one recorded trajectory segment reproduced nonlinear states, linearizations, gains and corrections, with agreement in the noise-factor covariance representation. Gaussian tests with fixed linearizations checked covariances without reconstructing means; six synthetic cases tested covariance stabilization, snapshot insertion, mapping and removal. These comparisons establish agreement for the specified numerical operations; the separate nonlinear calibration assessment is reported in the Discussion.

\subsection{Discrete propagation and updates}
\label{subsec:discrete}

First-order discretization gives
\begin{equation}
\mathbf{\Phi}_k \approx \mathbf{I} + \mathbf{F}_k \Delta t,
\end{equation}
\begin{equation}
\mathbf{P}_{k+1|k} = \mathbf{\Phi}_k \mathbf{P}_{k|k} \mathbf{\Phi}_k^\top + \mathbf{Q}_k,
\end{equation}
Here $\Delta t$ is the measured IMU interval and $\mathbf{Q}_k=\mathbf{Q}_c\Delta t$. The transition and reset use first-order linearizations of propagation and error injection.

\label{subsec:update}

For the DVL/depth prediction $h_{\mathrm{on}}$ and its right-error Jacobian $\mathbf{H}_k$ at the predicted nominal state, the innovation and gain are
\begin{equation}
\mathbf{r}_k = \mathbf{z}_k - h_{\mathrm{on}}\!\left(\hat{\mathbf{x}}_{k|k-1}\right),
\end{equation}
\begin{equation}
\mathbf{K}_k = \mathbf{P}_{k|k-1} \mathbf{H}_k^\top \left( \mathbf{H}_k \mathbf{P}_{k|k-1} \mathbf{H}_k^\top + \mathbf{R}_k \right)^{-1}.
\end{equation}

The onboard covariance uses the Joseph update before nominal-state error injection and the right-error reset.

\subsection{Computational cost and storage}
\label{subsec:complexity}

Compression refers to avoiding repeated trajectory passes while retaining explicit snapshot and source-pair covariance storage. Capacity $L$ allocates one $15\times15\times L\times L$ source-pair array, two $15\times15\times L$ covariance arrays, a $2\times L$ position array and $L$ packet IDs: $225L^2+(225+225+2+1)L=225L^2+453L$ doubles, or $8(225L^2+453L)$ bytes excluding MATLAB overhead. Current-to-source mapping is linear in active snapshots; removal copies the quadratic pair array. Retained $15\times15\times n$ diagnostic histories make total memory non-lag-bounded. The timing comparison includes full-pass execution and array handling in the evaluated Replay implementation.

\section{Experiments}
\label{sec:experiments}

\subsection{Communication-delay design}
\label{subsec:delay}

The primary study fixes delay at 1.5~s without outage; the separate prespecified sweep uses 0, 0.5, 1.5, 3, 5 and 10~s. Measurement realizations stay fixed; only scheduled reception changes. A delay of 1.5~s would correspond to 2.25~km one-way at 1500~m/s if entirely due to propagation, but neither range nor sound speed is simulated.

\subsection{Performance metrics}
\label{subsec:metrics}

Navigation performance is evaluated using the following metrics:
\begin{itemize}
  \item three-dimensional trajectory position root-mean-square error (RMSE), final position error, peak position error, and velocity RMSE;
  \item truth-supported 9-D normalized estimation error squared (NEES), summarized as average NEES (ANEES), for position, velocity, and attitude, together with DVL and acoustic normalized innovation squared (NIS) diagnostics; and
  \item mean timed core-loop computation per IMU step.
\end{itemize}
The NEES diagnostic uses position and velocity errors and the small-angle right-attitude error. Its 9-D covariance is symmetrized, with a pseudoinverse fallback if Cholesky factorization fails. The scalar archive lacks covariance matrices and fallback flags, so fallback counts and effective ranks are unknown.

Recordings form paired comparison blocks. Paired $t$ tests, Wilcoxon signed-rank tests, confidence intervals (CIs) and effect sizes describe this cohort; recording independence and nominal CI coverage have not been established. Runtime divides timed core-loop work by $n-1$ for direct filtering and $n-k_0$ for Replay ($n$: IMU samples; $k_0$: initialization index). Replay sums its initial pass and reception-triggered reruns, excluding wrapper scheduling/copying. Both timers exclude loading, measurement generation, pre-loop initialization/allocation and metrics. Reported runtimes retain these denominators; for a common per-IMU-step comparison, Replay times are rescaled by $(n-k_0)/(n-1)$ before averaging. The MAT header records PCWIN64, but not the original processor, OS, thread count or executing MATLAB release; R2024a source compatibility is not an execution record.

\subsection{Simulation setup}
\label{subsec:setup}
Of 200 candidate RexROV recordings, 46 lack DVL samples before filtering; 154 are input-usable. Whether production, logging or extraction caused the absence is unknown. IMU/ground-truth durations overlap, but start times differ (excluded 8.322--8.802~s; retained 32.662--51.802~s). Horizontal path ranges are 0.070--0.164 versus 54.597--108.232~m over full spans and 0.0079--0.0222 versus 13.252--28.662~m over the common 51.802--67.752~s interval. Thus, the retained cohort cannot be assumed representative of all 200. Methods share each usable input and noise realization, without performance-based censoring. Each retained recording has 401--404 processed acoustic updates; timing is normalized by IMU steps.

The sweep uses 30 paired input-usable recordings per delay with unchanged observations and noise. Its zero-delay gate checks the common sampled-time schedule, not exact alignment at $\tau_j$.

\subsubsection{Sensor and measurement generation}
Timestamps relative to the first IMU sample schedule updates; original clock/reset settings remain unverified. Nominal rates are 50~Hz (IMU), approximately 6.94~Hz (DVL) and 20~Hz (ground truth), but actual timestamps drive processing. Independent raw-bag decoding exactly reproduced all nine MAT fields for all 154 retained recordings, including times and quaternion order; this verifies stored content, not production identity or bias truth. Primary/sweep velocity is world-frame Odometry Twist copied without rotation and agrees with position derivatives, differing from the usual Robot Operating System (ROS) child-frame convention. The outage archive stores position without velocity/quaternion. Depth/acoustic fixes use linear ground-truth interpolation at DVL times $\tau_j$, with extrapolation beyond the recorded span. This defines synthetic means, not physical sub-sample truth or a stochastic bridge. Post-DVL snapshots at $t_{s,j}$ have observed $\delta_j$ of 0--0.020~s across the 154 inputs.

The matched 154-recording sensitivity at 1.5~s retained $\mathbf{p}_{xy}^{\mathrm{true}}(\tau_j)+\mathbf{v}_j^{\mathrm{ac}}$ in condition A and used $\mathbf{p}_{xy}^{\mathrm{true}}(t_{s,j})+\mathbf{v}_j^{\mathrm{ac}}$ in B, preserving noise, packet identity and timing. All A RMSEs reproduced the archive within $10^{-9}$~m. Mean B-minus-A RMSE was $+0.000194$~m (nominal 95\% percentile-bootstrap CI $[-0.000342,+0.000733]$~m; 10\,000 recording resamples); maximum absolute change was $0.009857$~m. Five extrapolated boundary candidates were generated but not processed. The interval assumes independent recordings; it does not verify population coverage, covariance conditioning or negligible timing effects.

Raw DVL is archived separately with unverified source noise. Added offline Gaussian perturbations have standard deviation $0.05$~m/s, matching the configured measurement setting; total recorded-plus-added noise is unknown. Depth ($0.10$~m) and acoustic ($5.0$~m/axis) draws use distinct segments of one pseudorandom stream; DVL uses another. Fixed $\mathbf{R}_{\mathrm{ac}}=25\mathbf{I}_{2}$~m$^2$ has no signal-to-noise dependence. All methods share the complete realization.

\subsection{Primary experiment results}
\label{subsec:stats}

\begin{figure}[t]
\centering
\includegraphics[width=8.5cm]{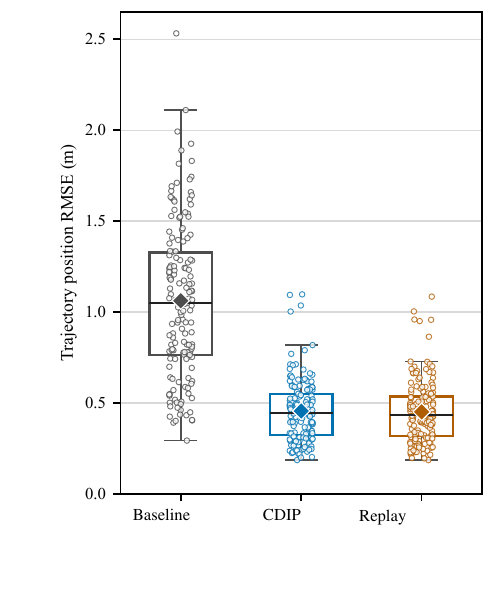}
\caption{Primary experiment using recorded simulation inputs and synthetic acoustic fixes (delay of 1.5~s; no outage): trajectory-position RMSE over ground-truth overlap (m). Open markers: all $N=154$ paired recordings/method; boxes: interquartile ranges and medians; whiskers: non-outlier range; diamonds: means.}
\label{fig:main_rmse_distribution}
\end{figure}

At the fixed acoustic delay of 1.5~s without an outage, all 462 method-recording runs across 154 input-usable paired recordings completed and were retained for analysis (Table~\ref{tab:main_performance}). Figure~\ref{fig:main_rmse_distribution} shows the recording-wise trajectory-RMSE distributions.

\begin{table}[htbp]
\centering
\caption{Primary paired experiment using recorded simulation inputs and synthetic acoustic fixes (delay of 1.5~s; no outage; $N=154$ per method). Errors over ground-truth overlap are mean $\pm$ standard deviation (units in headings); core-loop time per IMU step is a mean only. All runs completed.}
\label{tab:main_performance}
\small
\setlength{\tabcolsep}{2.5pt}
\begin{tabular}{lccccc}
\toprule
Method & \makecell{Traj.\\RMSE (m)} & \makecell{Final\\err. (m)} & \makecell{Peak\\err. (m)} & \makecell{Vel.\\RMSE (m/s)} & \makecell{Core-loop time\\(ms/IMU step)} \\
\midrule
Baseline & $1.062\pm0.421$ & $1.469\pm0.625$ & $1.825\pm0.682$ & $0.095\pm0.035$ & $0.0273$ \\
CDIP & $0.456\pm0.171$ & $0.346\pm0.220$ & $1.116\pm0.401$ & $0.057\pm0.020$ & $0.0743$ \\
Replay & $0.451\pm0.171$ & $0.347\pm0.218$ & $1.088\pm0.371$ & $0.057\pm0.019$ & $9.469$ \\
\bottomrule
\end{tabular}
\end{table}

CDIP's observed mean trajectory RMSE was $0.606$~m (57.1\%) below Baseline (nominal 95\% CI $[0.549,0.664]$~m; paired $t$ and Wilcoxon signed-rank tests, each $p<0.001$; Cohen's $d_z=1.68$). Final error, peak error and velocity RMSE means also favored CDIP (Table~\ref{tab:paired_main}). Sampling qualifications follow Section~\ref{subsec:metrics}.

\begin{figure}[t]
\centering
\includegraphics[width=8.5cm]{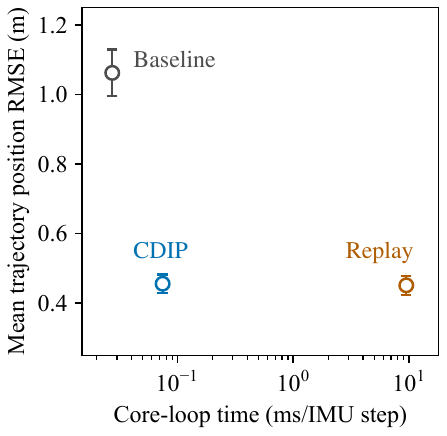}
\caption{Primary experiment using recorded simulation inputs and synthetic acoustic fixes (delay of 1.5~s; no outage): mean trajectory-position RMSE over ground-truth overlap (m) versus mean core-loop time (ms/IMU step). Points: saved means; bars: saved nominal 95\% CIs; runtime axis: logarithmic. The comparison describes the empirical accuracy--computation trade-off of the tested implementations.}
\label{fig:accuracy_runtime}
\end{figure}

CDIP approached Replay's mean trajectory RMSE, with a CDIP-minus-Replay difference of $+0.005$~m (nominal 95\% CI $[-0.001,0.010]$~m; paired $t$, $p=0.091$; Wilcoxon, $p=0.0187$; $d_z=0.137$). CDIP's mean RMSE was 1.03\% higher, with 99.2\% lower reported timed work per IMU step (Fig.~\ref{fig:accuracy_runtime}). Common-denominator adjustment changes the archived Replay/CDIP mean ratio from 127.394 to 127.355 without retiming. This comparison quantifies the empirical accuracy--computation trade-off for the tested implementations. Timing measures core-loop work; statistical equivalence, non-inferiority and end-to-end performance were outside the evaluated endpoints.

\begin{table}[htbp]
\centering
\caption{Paired primary comparisons using recorded simulation inputs and synthetic acoustic fixes (delay of 1.5~s; no outage; $N=154$ recordings). Trajectory-position RMSE over ground-truth overlap (m) is primary; other units appear by metric. Positive Baseline--CDIP and negative CDIP--Replay differences favor CDIP. Saved CIs and unadjusted paired $t$-test and Wilcoxon signed-rank $p$ values are nominal calculations under unverified recording-level sampling assumptions; secondary endpoint tests are descriptive. No multiplicity-adjusted $p$ values or equivalence/non-inferiority inference are claimed.}
\label{tab:paired_main}
\small
\setlength{\tabcolsep}{3.5pt}
\begin{tabular}{llcc@{\hspace{5pt}}c}
\toprule
Comparison & Metric & Mean difference (95\% CI) & $p$ ($t$/Wilcoxon) & Cohen's $d_z$ \\
\midrule
Baseline--CDIP & Traj. RMSE (m) & $0.606$ $[0.549,0.664]$ & $<0.001/<0.001$ & $1.68$ \\
Baseline--CDIP & Final err. (m) & $1.123$ $[1.025,1.222]$ & $<0.001/<0.001$ & $1.81$ \\
Baseline--CDIP & Peak err. (m) & $0.709$ $[0.616,0.802]$ & $<0.001/<0.001$ & $1.21$ \\
Baseline--CDIP & Vel. RMSE (m/s) & $0.038$ $[0.033,0.042]$ & $<0.001/<0.001$ & $1.34$ \\
\midrule
CDIP--Replay & Traj. RMSE (m) & $0.005$ $[-0.001,0.010]$ & $0.0912/0.0187$ & $0.137$ \\
CDIP--Replay & Final err. (m) & $-0.001$ $[-0.005,0.003]$ & $0.615/0.906$ & $-0.041$ \\
CDIP--Replay & Peak err. (m) & $0.028$ $[0.007,0.048]$ & $0.00834/<0.001$ & $0.215$ \\
CDIP--Replay & Vel. RMSE (m/s) & $0.001$ $[-0.000,0.001]$ & $0.0927/0.0901$ & $0.136$ \\
CDIP--Replay & \makecell[l]{Core-loop time\\(ms/IMU step)} & $-9.394$ $[-9.649,-9.140]$ & $<0.001/<0.001$ & $-5.88$ \\
\bottomrule
\end{tabular}
\end{table}

\subsection{Consistency and numerical integrity}
\label{subsec:consistency}

Position/velocity/attitude ANEES uses valid recordings at 60 times: 49 at the first (95\% working bounds $[7.851,10.226]$), then 154 at the remaining 59 ($[8.342,9.682]$). Baseline was predominantly above the upper bound; CDIP and Replay were below the lower bound at half the times, consistent with conservative covariance behavior relative to these working references. Interpretation is descriptive because independent-Gaussian, covariance and rank assumptions remain unverified.

\begin{table}[htbp]
\centering
\caption{Primary experiment using recorded simulation inputs and synthetic acoustic fixes (delay of 1.5~s; no outage): dimensionless 9-D NEES summaries and DVL/acoustic NIS over processed measurements. NEES means first average over each recording's valid times, then over recordings. Fractions classify pointwise ANEES against 95\% working reference bounds, using 49 valid recordings first and 154 thereafter. Expected NIS dimensions are three (DVL) and two (acoustic).}
\label{tab:consistency}
\small
\setlength{\tabcolsep}{4pt}
\begin{tabular}{lcccc}
\toprule
Method & \makecell{Mean time-averaged\\9-D NEES} & \makecell{Below / inside\\/ above} & \makecell{Mean DVL\\NIS} & \makecell{Mean acoustic\\NIS} \\
\midrule
Baseline & $18.110$ & $0.0\% / 11.7\% / 88.3\%$ & $2.995$ & $2.084$ \\
CDIP & $8.728$ & $50.0\% / 40.0\% / 10.0\%$ & $2.990$ & $1.993$ \\
Replay & $8.691$ & $50.0\% / 40.0\% / 10.0\%$ & $2.990$ & $1.992$ \\
\bottomrule
\end{tabular}
\end{table}

\begin{figure*}[t]
\centering
\includegraphics[width=17.0cm]{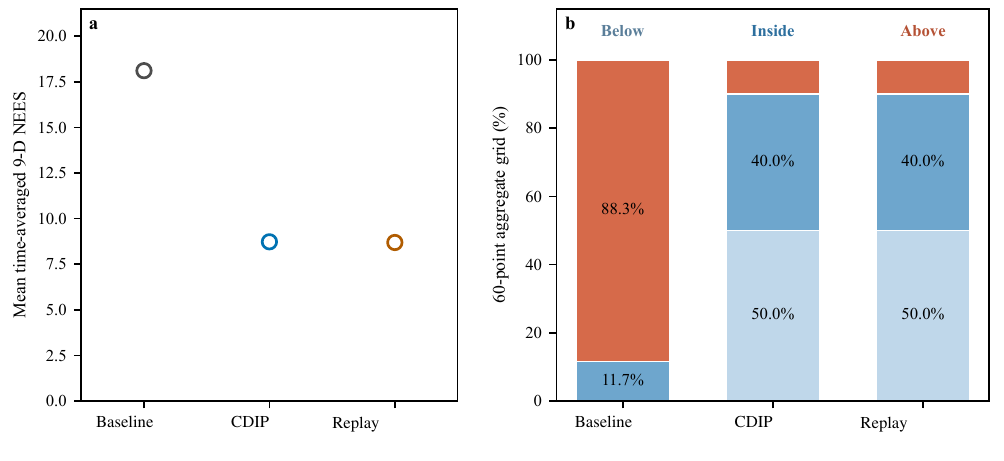}
\caption{Primary experiment using recorded simulation inputs and synthetic acoustic fixes (delay of 1.5~s; no outage): dimensionless truth-supported 9-D uncertainty diagnostics over 60 common evaluation times. (a) Mean time-averaged 9-D NEES using the aggregation in Table~\ref{tab:consistency}; no confidence interval is shown. (b) Fractions of the grid below, inside, and above the 95\% pointwise ANEES working bounds calculated from each time's valid-recording count. Grid fractions describe pointwise classifications rather than independent-sample coverage. Both panels provide descriptive diagnostics for the nine-dimensional position, velocity and attitude error.}
\label{fig:consistency_summary}
\end{figure*}

Mean NIS was near the DVL/acoustic reference dimensions (Table~\ref{tab:consistency}; Fig.~\ref{fig:consistency_summary}), without testing innovation centering, whiteness or covariance calibration. Stabilized current covariance had minimum eigenvalue $>10^{-9}$ and zero asymmetry; quaternion-norm error was $<10^{-15}$. These checks describe current-state numerical integrity; nonlinear calibration is assessed separately in the Discussion.

\subsection{Separate fixed-delay sweep}
\label{subsec:delay_sweep_results}

All 90 zero-delay pairwise checks passed on the shared IMU grid rather than exact generation time. All sweep runs completed and were retained (Table~\ref{tab:delay_sweep}).

\begin{table}[htbp]
\centering
\caption{Fixed-delay sweep using recorded simulation inputs and synthetic acoustic fixes ($N=30$ paired recordings/delay). Entries are mean trajectory RMSE over ground-truth overlap (m) and mean core-loop time (ms/IMU step). CDIP minus Replay RMSE differences have nominal 95\% CIs at nonzero delays; negative values favor CDIP. All intervals include zero.}
\label{tab:delay_sweep}
\small
\setlength{\tabcolsep}{3.5pt}
\begin{tabular}{rcccccc}
\toprule
\makecell{Delay\\(s)} & \makecell{Baseline\\RMSE} & \makecell{CDIP\\RMSE} & \makecell{Replay\\RMSE} & \makecell{CDIP--Replay\\(95\% CI)} & \makecell{CDIP\\runtime} & \makecell{Replay\\runtime} \\
\midrule
$0$ & $0.455$ & $0.455$ & $0.455$ & --- & $0.0307$ & $9.565$ \\
$0.5$ & $0.535$ & $0.457$ & $0.462$ & $-0.005$ $[-0.015,0.005]$ & $0.0416$ & $9.207$ \\
$1.5$ & $1.053$ & $0.469$ & $0.477$ & $-0.008$ $[-0.023,0.006]$ & $0.0877$ & $10.774$ \\
$3$ & $1.934$ & $0.511$ & $0.501$ & $0.010$ $[-0.007,0.027]$ & $0.1141$ & $7.980$ \\
$5$ & $2.948$ & $0.531$ & $0.537$ & $-0.006$ $[-0.035,0.023]$ & $0.2409$ & $7.664$ \\
$10$ & $4.354$ & $0.674$ & $0.631$ & $0.043$ $[-0.008,0.094]$ & $0.6525$ & $6.834$ \\
\bottomrule
\end{tabular}
\end{table}

Baseline mean trajectory RMSE rose from $0.455$~m at zero delay to $4.354$~m at 10~s (Table~\ref{tab:delay_sweep}). CDIP and Replay reached at most $0.674$ and $0.631$~m, respectively; their mean ordering changed across delays, and every nonzero-delay paired difference CI included zero. Across the sweep, the relative mean RMSE therefore depended on delay.

\begin{figure*}[t]
\centering
\includegraphics[width=17.0cm]{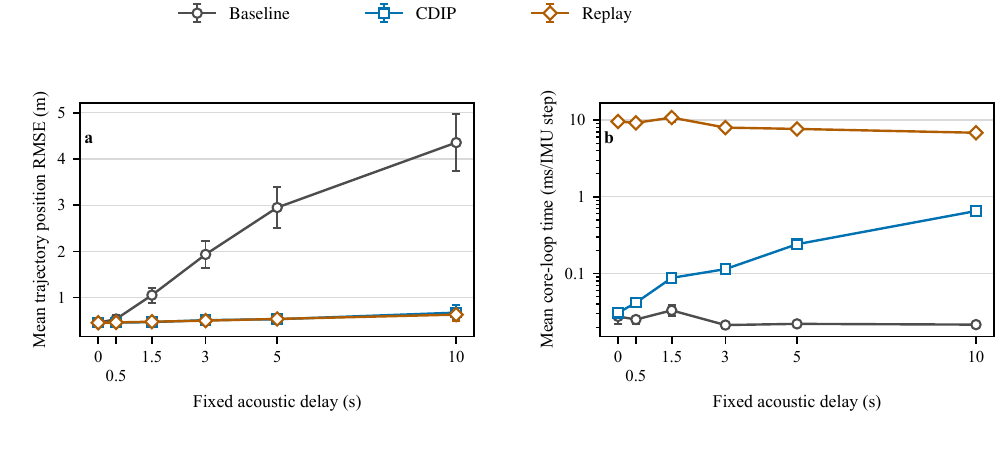}
\caption{Fixed-delay sweep using recorded simulation inputs and synthetic acoustic fixes ($N=30$ paired recordings/delay). (a) Mean trajectory-position RMSE over ground-truth overlap (m) and saved nominal 95\% CIs. (b) Mean core-loop time (ms/IMU step) and saved nominal 95\% CIs; log vertical axis. Replay runtime decreases with delay as fewer packets arrive within the fixed evaluable horizon, not because per-packet Replay is cheaper.}
\label{fig:delay_accuracy_runtime}
\end{figure*}

CDIP's mean core-loop time rose from $0.0416$~ms/IMU step at 0.5~s to $0.6525$~ms/IMU step at 10~s as cross-covariances were retained longer; the measured full-from-start Replay implementation remained slower throughout (Fig.~\ref{fig:delay_accuracy_runtime}).

\subsection{Synthetic acoustic packet-erasure sensitivity}
\label{subsec:packet_erasure_sensitivity}

The erasure study uses the first 30 input-usable recordings, separately from the primary cohort. At 1.5~s, methods share one mask per recording with 10\% erasure nested within 30\%. Only acoustic availability changes; observation values, covariance, reception times and non-acoustic inputs stay fixed. All 270 method--recording--rate RMSEs were finite, including 90 at zero erasure; 180 nonzero rows passed saved numerical checks (Table~\ref{tab:packet_erasure_sensitivity}).

\begin{table}[htbp]
\centering
\caption{Packet-erasure sensitivity using recorded simulation inputs and synthetic acoustic fixes (delay of 1.5~s; 30 paired recordings): mean trajectory-position RMSE over ground-truth overlap (m). One nested mask realization per recording is shared across methods.}
\label{tab:packet_erasure_sensitivity}
\small
\setlength{\tabcolsep}{3pt}
\begin{tabular}{lccc}
\toprule
Method & 0\% & 10\% & 30\% \\
\midrule
Baseline & $1.052583$ & $1.009340$ & $0.973824$ \\
CDIP & $0.468773$ & $0.479721$ & $0.481825$ \\
Replay & $0.477181$ & $0.489459$ & $0.499056$ \\
\bottomrule
\end{tabular}
\end{table}

Each CDIP RMSE at 10\% or 30\% erasure was paired with its zero-erasure value. Mean changes were $+0.010947861$~m (unadjusted 95\% paired-$t$ CI $[-0.006861205,0.028756927]$~m; two-sided $p=0.218685677$) and $+0.013052230$~m (CI $[-0.018982775,0.045087234]$~m; $p=0.411485830$). Two-test Holm adjustment gives $p_{\mathrm{Holm}}=0.437371353$ for each; intervals are not multiplicity-adjusted. Both adjusted p values exceed the nominal familywise 0.05 threshold, but unverified recording independence prevents population effect or no-degradation inference. Baseline's declining mean RMSE has no established cause (Table~\ref{tab:packet_erasure_sensitivity}). Exploratory checks suggested upper-tail non-normality at 10\%; no seed was excluded. The paired summaries describe this nonrandom cohort and its single-mask realizations; equivalence and physical-channel robustness require additional evidence.

\subsection{Robustness under prolonged acoustic outages}
\label{subsec:outage_robustness}

The targeted stress test uses five paired scenarios for each outage of 50, 100, 200 and 300~s at a fixed delay of 1.5~s. Outage packets are permanently lost, without batch recovery. All 60 method-scenario runs completed (Table~\ref{tab:outage_robustness}).

\begin{table}[htbp]
\centering
\caption{Permanent-loss outage study using recorded simulation inputs and synthetic acoustic fixes (delay of 1.5~s; five paired scenarios/duration). Entries are mean $\pm$ sample standard deviation of trajectory-position RMSE over ground-truth overlap (m); CDIP reduction is relative to the Baseline mean.}
\label{tab:outage_robustness}
\small
\setlength{\tabcolsep}{3.5pt}
\begin{tabular}{rcccc}
\toprule
\makecell{Outage\\(s)} & \makecell{Baseline} & \makecell{CDIP} & \makecell{Replay} & \makecell{CDIP\\reduction} \\
\midrule
$50$  & $1.816\pm0.048$ & $0.412\pm0.020$ & $0.421\pm0.028$ & $77.3\%$ \\
$100$ & $1.825\pm0.035$ & $0.412\pm0.016$ & $0.421\pm0.023$ & $77.4\%$ \\
$200$ & $1.826\pm0.071$ & $0.435\pm0.041$ & $0.443\pm0.045$ & $76.2\%$ \\
$300$ & $1.920\pm0.217$ & $0.470\pm0.041$ & $0.478\pm0.044$ & $75.5\%$ \\
\bottomrule
\end{tabular}
\end{table}

\begin{figure*}[t]
\centering
\includegraphics[width=14.0cm]{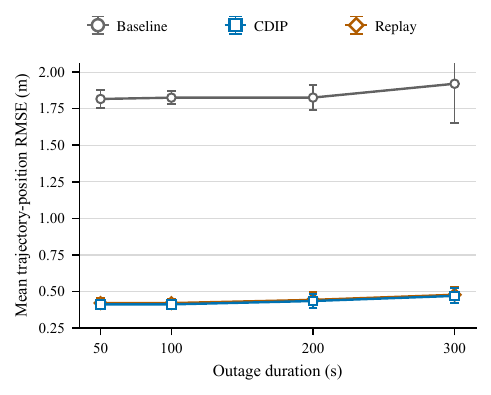}
\caption{Outage study using recorded simulation inputs and synthetic acoustic fixes (delay of 1.5~s; $n=5$ paired scenarios/duration). Points: mean trajectory-position RMSE over ground-truth overlap (m); bars: saved nominal 95\% CIs. Outage packets were permanently lost without batch recovery.}
\label{fig:outage_robustness}
\end{figure*}

Across 50--300~s outages, CDIP reduced mean trajectory RMSE by 75.5--77.4\% against Baseline (Table~\ref{tab:outage_robustness}; Fig.~\ref{fig:outage_robustness}). Gaps from Replay means were 7.7--9.1~mm; all nominal paired CDIP-minus-Replay 95\% CIs included zero ($t$ tests, $p=0.100$--$0.113$). These intervals summarize paired differences for the tested scenarios.

Mean core-loop time was $0.150$--$0.165$~ms/IMU step for CDIP and $325$--$456$~ms/IMU step for Replay. These stress-test timings are separate from the primary 99.2\% contrast.

\section{Discussion}
\label{sec:discussion}

CDIP's observed accuracy--computation trade-off supports explicit treatment of historical observations: it approached Replay's mean accuracy with lower core-loop computation. Fixed delay and noise isolate packet treatment; the sweep varies measurement age, and outage/erasure studies vary availability. The comparison evaluates complete implementations; attributing improvements to individual covariance maps would require component ablation.

Stochastic cloning addresses relative-state measurements and exact OOSM addresses tracking \cite{RoumeliotisBurdick2002Cloning,Mourikis2007SCKF,BarShalom2002Exact}. VHD synthesizes outage-time virtual measurements, whereas the two-speed method addresses cooperative navigation \cite{li2026communicationoutageresistantuuvstate,li2026asynchronous}. Their different states and aiding protocols prevent a matched comparison here.

The computational comparison uses full-from-start Replay; physical-channel effects and optimized fixed-lag methods lie outside the timestamped-fix study. Baseline RMSE also decreased under erasure, a pattern requiring further investigation.

During permanent loss, CDIP outage-period RMSE increased from $0.332$~m at 50~s to $0.620$~m at 300~s; post-outage means were $0.379$--$0.525$~m versus Baseline's $1.746$--$1.850$~m. These measurements characterize the tested permanent-loss use case, with lost packets remaining unrecovered.

\subsection{Limitations and threats to validity}

The simulation model isolates delayed-packet handling; range-dependent propagation, multipath, modem/network scheduling and clock error lie outside its scope. CDIP's frozen-history approximation uses $t_{s,j}$ for observations generated at $\tau_j$, omitting temporal error and associated conditional variance/cross-covariances. The observed 0--0.020~s offsets and observation-mean sensitivity characterize sampled timing mismatch; its full covariance implications remain unquantified. Full conditioning would require a common information set and source/current/cross-source correlations. The 154-recording diagnostics are descriptive under working sampling assumptions; a 15-D calibration analysis also requires unavailable bias truth.

An additional assessment used 256 independently generated trials of a separate nonlinear synthetic model, paired across estimators and aligned or 10/20-ms offset conditions. Neither CDIP nor the two fully augmented ESKF references met the prespecified calibration criterion. Among 360 prespecified tests, 72 NEES mean/coverage tests showed significant departures after Holm correction (familywise $\alpha=0.05$), covering both 9-D and 15-D errors, including aligned augmented references. At 12~s, mean 15-D NEES ranged from 59.99 to 62.89 across six estimator/condition groups. NEES reconstruction and packet-time indexing checks reproduced the results; the statistical mismatch cannot be isolated to frozen history or generation/snapshot offsets because departures also occur in fully augmented references. Uncertainty statistics are therefore reported descriptively rather than as calibrated coverage bounds.

The 154 input-usable recordings define the evaluation cohort; the other 46 differ in motion and start time. Five outage pairs per duration and one erasure mask per recording bound the availability tests. Exact bag/MAT correspondence and exploratory gyro-frame support were checked; full IMU mounting rotation/lever arm, physical bias/noise, production configuration and seed/reset require separate verification. Original timing-host metadata were not archived. Field tests, optimized fixed-lag comparisons, and adaptation to raw ranges and uncertain clocks remain future work.

Future tests could couple CDIP to network simulations, including Aqua-Sim Fourth Generation \cite{Guo2025AquaSimFG}, and examine distributed multi-UUV navigation, stochastic loss/recovery, clock error, and field data.

\section{Conclusion}
\label{sec:conclusion}

CDIP incorporates delayed horizontal acoustic-position fixes into a causal single-UUV ESKF through source-associated IMU snapshots and maintained cross-covariances. Its frozen-history projection corrects the current state while avoiding repeated full-from-start filtering.

In 154 paired simulation recordings at a delay of 1.5~s without outage, CDIP reduced mean trajectory-position RMSE by 57.1\% against Baseline, achieving 0.456~m compared with Replay's 0.451~m. Mean core-loop time per processed IMU step was 99.2\% lower than for the tested full-from-start Replay implementation. The six-delay sweep, 50--300~s scheduled permanent-loss outages and separate 30-recording packet-erasure study described performance across measurement-timing and availability conditions.

\Needspace{3\baselineskip}

The 9-D diagnostics describe covariance behavior relative to working bounds. A separate nonlinear synthetic assessment found calibration departures in CDIP and fully augmented ESKF references, as discussed above; further uncertainty calibration is a next step. Within the evaluated simulations, CDIP provides an explicitly approximate route to near-Replay mean positioning accuracy with substantially lower measured core-loop computation.

\begin{acknowledgments}
This research was partially funded by Postgraduate Research Scholarship (PGRS) at Xi’an Jiaotong-Liverpool University (FOS2312JBD01), Suzhou Municipal Key Laboratory Broadband Wireless Access Technology (BWAT) and JITRI Supervision Support Fund (JSF10120220008) of XJTLU-JITRI Academy.

\end{acknowledgments}

\section*{Author Declarations}
\subsection*{Conflict of Interest}
The authors have no conflicts to disclose.

\section*{Data Availability}
The data supporting the reported primary, delay-sweep, outage, packet-erasure, and separate synthetic-calibration analyses are available in Zenodo dataset version 2.0 at \url{https://doi.org/10.5281/zenodo.23795814} \cite{Li2026CDIPDataset}. The release includes the unchanged original 18-CSV results archive and an additional validation archive. Raw ROS recordings, complete original simulation inputs and result containers, MATLAB estimator implementation code, and complete synthetic packet-time arrays are not included. Consequently, the deposit supports inspection and reanalysis of saved results but not complete end-to-end executable reproduction.

\makeatletter
\auto@bib@empty
\begingroup
\def\thebibliography#1{\savethebibliography{#1}\advance\parskip1pt\relax}
\bibliography{ref}
\endgroup
\makeatother

\end{document}